\documentclass[10pt,twocolumn,letterpaper]{article}

\usepackage{arxiv}
\usepackage{times}
\usepackage{epsfig}
\usepackage{graphicx}
\usepackage{amsmath}
\usepackage{amssymb}
\usepackage{multirow}
\usepackage{diagbox}
\usepackage{caption}
\usepackage{subfigure}
\usepackage{color}
\usepackage{bm}
\usepackage{algorithm}

\usepackage[breaklinks=true,bookmarks=false]{hyperref}

\cvprfinalcopy 

\begin{document}
	
	\title{Layer-Wise Adaptive Updating for Few-Shot Image Classification}

	\author{Yunxiao Qin, Weiguo Zhang, Zezheng Wang, Chenxu Zhao, Jingping Shi \\
	qyxqyx@mail.nwpu.edu.cn, zhangwg@nwpu.edu.cn, zezhengwang@aibee.com \\ zhaochenxu@mininglamp.com, shijingping@nwpu.edu.cn
}

	\maketitle
	\begin{abstract}
		Few-shot image classification (FSIC), which requires a model to recognize new categories via learning from few images of these categories, has attracted lots of attention. Recently, meta-learning based methods have been shown as a promising direction for FSIC. Commonly, they train a meta-learner (meta-learning model) to learn easy fine-tuning weight, and when solving an FSIC task, the meta-learner efficiently fine-tunes itself to a task-specific model by updating itself on few images of the task. In this paper, we propose a novel meta-learning based layer-wise adaptive updating (LWAU) method for FSIC. LWAU is inspired by an interesting finding that compared with common deep models, the meta-learner pays much more attention to update its top layer when learning from few images. According to this finding, we assume that the meta-learner may greatly prefer updating its top layer to updating its bottom layers for better FSIC performance. Therefore, in LWAU, the meta-learner is trained to learn not only the easy fine-tuning model but also its favorite layer-wise adaptive updating rule to improve its learning efficiency. Extensive experiments show that with the layer-wise adaptive updating rule, the proposed LWAU: 1) outperforms existing few-shot classification methods with a clear margin; 2) learns from few images more efficiently by at least 5 times than existing meta-learners when solving FSIC.

	\end{abstract}
	
	\section{Introduction}
	\label{sec:intro}
	Deep learning based artificial intelligence has shown remarkable progresses in many computer vision tasks\cite{image-classification1, image-classification2, dong2017multiscale,Tcnn, image-classification3, machine-trans1}.
	However, it still extremely falls behind human intelligence in the aspect of learning from few images.
	Few-shot image classification (FSIC)\cite{6842695,match-network, prototypical,MAML,SNAIL}, which aims deep models to recognize unseen categories by learning from few images of these categories, becomes a more and more popular problem.
	
	Some researchers have tried to solve the FSIC problem via metric learning\cite{huang2020class-prototype,match-network, prototypical,jiang2019adaptive}. 
	The principal of these approaches\cite{match-network, prototypical} is to train a non-linear mapping function that represents images into an embedding space. 
	After training, the embeddings of images belonging to different classes are easy to be distinguished by classifying the embeddings using the nearest neighbor\cite{Tknn,TKNN2} or linear classifiers.
	
	Recently, meta-learning based FSIC methods \cite{TAML, On-the-optimiazation, An-alternative-to, NIPS2019_9468, Sun_2019_CVPR, MAML,Memory_augment,SNAIL,RL2, Achille_2019_ICCV, Lee_2019_CVPR, zhuang, Metagan} are more popular.
	Instead of training a model on images, they commonly train a meta-learner on FSIC tasks for the meta-learner learning a universe easy fine-tuning initial weight for FSIC tasks\cite{MAML,iMAML,TAML,qin2020learning}.
	In each FSIC task, the meta-learner is required to recognize new image categories by fine-tuning itself based on few images of these categories.
	Typically, FSIC task is called \emph{N}-way \emph{K}-shot classification task\cite{match-network}, and the task is constructed with a support set for learners (common deep models or meta-learners) fine-tuning and a query set for evaluation. 
	\emph{N}-way means the task contains \emph{N} unseen categories for the learner to recognize and \emph{K}-shot means the support set contains \emph{K} samples for each of the category for the learner to fine-tune.
	
	
	
	\begin{figure}[t]
		\centering
		\includegraphics[width=0.35\textwidth]{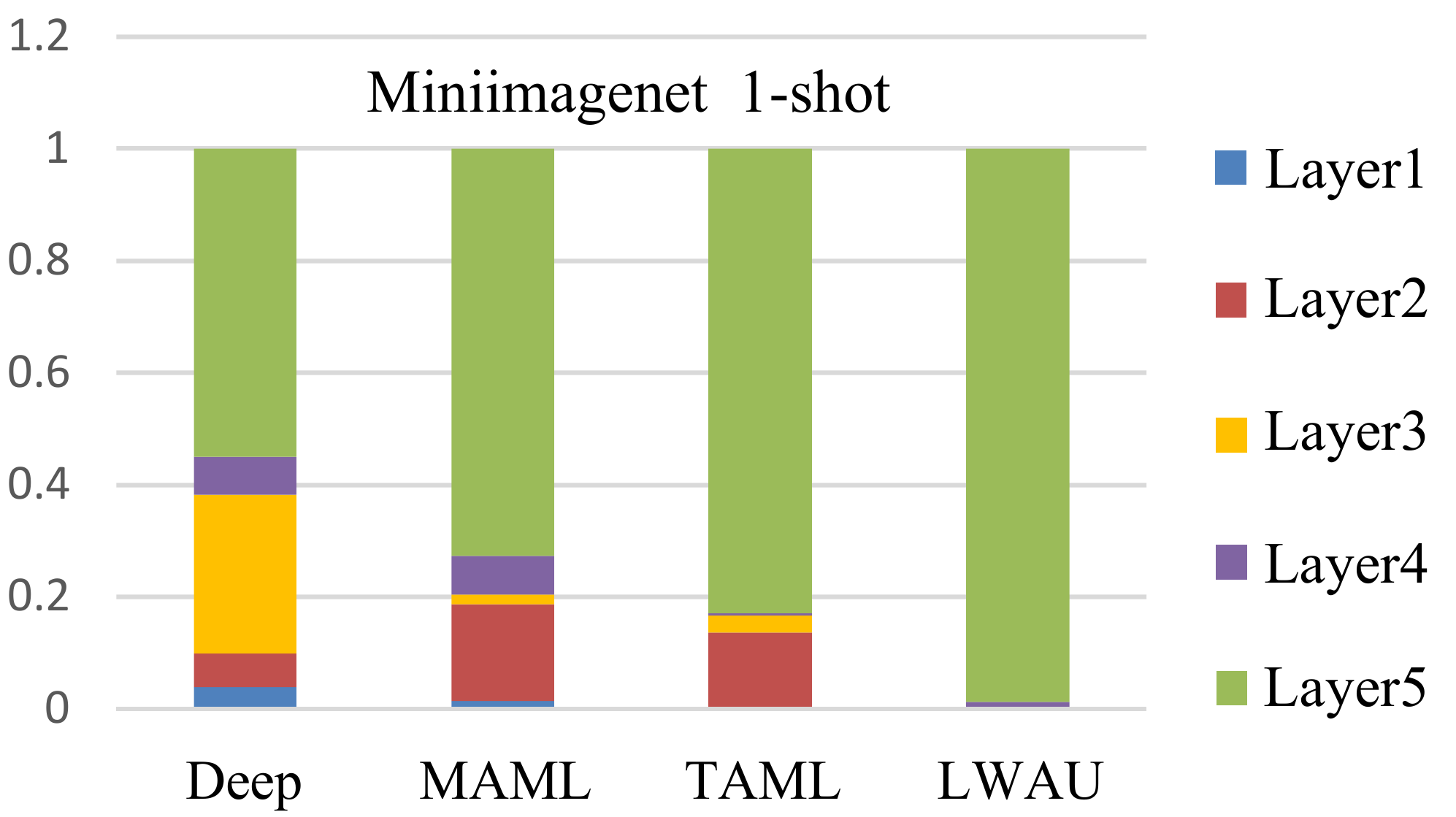}
		\caption{
			The update proportions of network layers of the deep model, MAML\cite{MAML}, TAML\cite{TAML} and LWAU.
			The proportion of each layer is calculated with Eq.\ref{eq:proportion} when the model/meta-learner fine-tuning itself on the support set of 600 5-way 1-shot testing tasks generated on Miniimagenet.
			All their network is the Conv-4 network which consists of four cascaded convolution layers and one fully-connect layer.
			``Layer5" denotes the top fully-connect layer and 'Layers1-4' denote the convolution layers.
		}
		\label{fig:intro}
	\end{figure}
	
	In this paper, we discover an interesting phenomenon of these meta-learning based methods. 
	As shown in Fig.\ref{fig:intro}, when fine-tuning on the support set, compared with deep model, the meta-learners pay more attention to update its top layer.
	For example, the update proportions of the top (5-th) layer of the MAML\cite{MAML} and TAML\cite{TAML} are approximately 73\% and 83\%, while that of the deep model is 55\%.
	For easily understanding, we formulate the update proportion as
	\begin{equation}
	p_i = \frac{\|\Delta w_i/w_i\|_2 }{\sum^n_{j}\|\Delta w_j/w_j\|_2},
	\label{eq:proportion}
	\end{equation}
	where $p_i$ denotes the update proportion of the $i$-th layer, $w_i$ and $w_i + \Delta w_i$ represent the initial weight and updated weight after $L$ (10 in this work) update steps of the $i$-th layer, respectively.
	$\|*\|_2$ is L2 norm and $n$ denotes the number of layers.
	Note that, in this paper, update denote the meaning that the meta-learner fine-tunes its weight on the support set of a FSIC task.
	
	Inspired by the above phenomenon, we make a study on the meta-learning based FSIC and contribute to the few-shot image classification community in three folds.
	
	1) We assume that in the FSIC scene, the meta-learner may greatly prefer updating its top layer to updating its bottom layers for better performance.
	In other words, a better-suited layer-wise updating rule may is favored by the meta-learner. 
	To validate the assumption and improve the meta-learner's FSIC performance, we design a novel layer-wise adaptive updating (LWAU) method\footnotemark[1].
	Manually designing the better-suited layer-wise updating rule is inefficient and expensive.
	So, in LWAU, we train the meta-learner to learn not only the easy fine-tuning weight but also its favorite layer-wise adaptive updating rule.
	
	\footnotetext[1]{Code is available at https://github.com/qyxqyx/LWAU.}
	
	2) Extensive experiments conducted on two FSIC benchmarks (\emph{e.g.}, Miniimagenet, and Tieredimagenet) validate the assumption and LWAU. 
	As shown in Fig.\ref{fig:intro}, when the LWAU meta-learner updating itself on the support set of FSIC tasks, it almost neglects its bottom layers and pays almost all attention to update its top layer.
	Meanwhile, the proposed LWAU apparently outperforms the other few-shot classification methods on both Miniimagenet and Tieredimagenet, which shows the effectiveness of the layer-wise updating rule.
	Besides, we visualize the learned sparse image representations of LWAU for a better understanding of LWAU.
	
	3) We show that when tested on FSIC tasks, the proposed LWAU meta-learner can accelerate its update by updating only its top layers without performance decline.
	For example, compared with traditional meta-learning based methods which need updating all layers, LWAU can speed up the update for 5 and 10 times on 1- and 5- shot tasks on Miniimagenet, respectively.

	\section{Methodology}
\begin{figure}[t]
		\centering
		\includegraphics[width=0.32\textwidth]{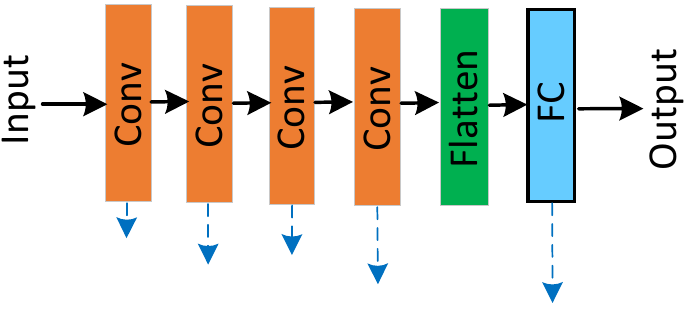}
		\caption{
			The Conv-4 network.
			It consists of four cascaded convolution layers to extract features and one fully-connect layer to predict.
			The different lengths of blue dotted arrows denote the update proportions of different layers.
		}
		\label{fig:fig2}
	\end{figure}

	In this section, we detail the proposed layer-wise adaptive updating (LWAU) method.
	For a fair comparison with the other few-shot classification methods, the network used in LWAU is the same as that in \cite{match-network}. 
	We call the network Conv-4 and show its structure in Fig.\ref{fig:fig2}. 
	It consists of five layers including four cascaded convolutions and one fully-connect.
	On an FSIC task $\tau_k$, we train the meta-learner with the following three stages.
	
	\textbf{First}, the meta-learner updates its weight on the support set for recognizing the categories in task $\tau_k$.
	Note that, each layer updates with an exclusive learning-rate rather then the global learning-rate shared with the other layers.
	For clarity, we show only one update step of LWAU, and the update can be formulated as Eq.\ref{eq:support_update1} and Eq.\ref{eq:support_update2}.
	
	\begin{equation}
	\emph{L}_{s(\tau_k)}(\bm{\theta}) \leftarrow \frac{1}{N_s(\tau_k)} \displaystyle{\sum_{x,y \in s(\tau_k)}^{}}\emph{l}(\emph{f}_{\bm{\theta}}(x),y) 
	\label{eq:support_update1}
	\end{equation}
	\begin{equation}
	\bm{\theta^{'}} \leftarrow \bm{\theta}-\bm{\alpha}{\boldmath \cdot}\nabla_{\bm{\theta}}\emph{L}_{s(\tau_k)}(\bm{\theta})
	\label{eq:support_update2}
	\end{equation}
	
	$\bm{\theta} = [\theta_1, \theta_2, \theta_3, \theta_4, \theta_5]$ is the weight vector of the five layers (\emph{i.e.}, $\theta_i$ is the weight of $i$-th layer, where $i\in$ [1, 5]).
	$\bm{\alpha} = [\alpha_1, \alpha_2, \alpha_3, \alpha_4, \alpha_5]$ is a vector and each $\alpha_i\in\bm{\alpha}$ is a trainable scaler denoting the exclusive updating learning-rate of $\theta_i$.
	$s(\tau_k)$ is the support set of the task $\tau_k$, and $x$ and $y$ is a pair of instance and label belonging to the support set of $\tau_k$. 
	$l$ is the cross-entropy classification loss function.
	$\emph{L}_{s(\tau_k)}(\bm{\theta})$ is the meta-learner's loss on the support set, and $\emph{f}_{\bm{\theta}}(x)$ is the prediction of the meta-learner.
	$N_s(\tau_k)$ denotes the number of instances in the support set.
	With the update on the support set, each layer's weight $\theta_i$ turns to $\theta_i^{'}$ with its exclusive updating learning-rate $\alpha_i$. 
	
	\begin{algorithm}[t]
		\caption{LWAU for Few-Shot Classification}
		{\bfseries input:} Few-shot classification training task list \emph{$T$}, learning-rate $\beta$, number of ways $N$, number of shots $K$. \\
		{\bfseries output:} Weights $\bm{\theta}$ and layer-wise updating learning-rate $\bm{\alpha}$ \\
		{\bfseries 1 \;\!:} initialize $\bm{\theta}$ and $\bm{\alpha}$ \\
		{\bfseries 2 \;\!:} {\bfseries while not} done {\bfseries do} \\
		{\bfseries 3 \;\!:} 	\quad	sample batch tasks $\tau$$_i$ $\in$ \emph{$T$}\\
		{\bfseries 4 \;\!:}  \quad   {\bfseries for} each of  $\tau$$_i$  {\bfseries do} \\
		{\bfseries 5 \;\!:} \qquad	$\emph{L}_{s(\tau_k)}(\bm{\theta}) \leftarrow \frac{1}{N_s(\tau_k)} \displaystyle{\sum_{x,y \in s(\tau_k)}^{}}\emph{l}(\emph{f}_{\bm{\theta}}(x),y)$   \\
		{\bfseries 6 \;\!:} 	\qquad  $\bm{\theta^{'}} \leftarrow \bm{\theta}-\bm{\alpha}{\boldmath \cdot}\nabla_{\bm{\theta}}\emph{L}_{s(\tau_k)}(\bm{\theta})$\\
		{\bfseries 7 \;\!:} 	\qquad	$\emph{L}_{q(\tau_k)}(\bm{\theta^{'}}) \leftarrow \frac{1}{N_q(\tau_k)} \displaystyle{\sum_{x,y \in q(\tau_k)}^{}}\emph{l}(\emph{f}_{\bm{\theta^{'}}}(x),y)$ \\
		{\bfseries 8 \;\!:} 	\quad {\bfseries end} \\
		{\bfseries 9 \;\!:} 	\quad $\bm{\theta}, \bm{\alpha} \leftarrow \bm{\theta}, \bm{\alpha} - \beta\cdot\nabla_{\bm{\theta}, \bm{\alpha}}\sum_{\tau_k}^{}\emph{L}_{q(\tau_k)}(\bm{\theta^{'}})$ \\
		{\bfseries 10:} 	{\bfseries end}
		\label{algorithm:train}
	\end{algorithm}

	\textbf{Secondly}, the meta-learner with its updated weight is evaluated on the query set, which can be formulated as Eq.\ref{eq:query_update1}.
	\begin{equation}
	\emph{L}_{q(\tau_k)}(\bm{\theta^{'}}) \leftarrow \frac{1}{N_q(\tau_k)} \displaystyle{\sum_{x,y \in q(\tau_k)}^{}}\emph{l}(\emph{f}_{\bm{\theta^{'}}}(x),y)
	\label{eq:query_update1}
	\end{equation}
	
	$q(\tau_k)$ is the query set of the task $\tau_k$, and 
	$N_q(\tau_k)$ is the number of instances in the query set and $\emph{L}_{q(\tau_k)}(\bm{\theta^{'}})$ is the meta-learner's loss on the query set.
	Note that when evaluating the meta-learner on the query set, the meta-learner predicts the instance $x$ with the updated weight $\bm{\theta^{'}}$. 
	
	\textbf{Finally}, for the meta-learner learning the easy fine-tuning weight $\bm{\theta}$ and its preferable layer-wise adaptive learning rule $\bm{\alpha}$ so that the meta-learner can fine-tune itself precisely on the support set to recognize categories of the task, both $\bm{\theta}$ and $\bm{\alpha}$ are meta-trained.
	Thus, the training of LWAU is
	\begin{equation}
	(\bm{\theta}, \bm{\alpha}) \leftarrow (\bm{\theta}, \bm{\alpha}) - \beta\cdot\nabla_{(\bm{\theta}, \bm{\alpha})}\emph{L}_{q(\tau_k)}(\bm{\theta^{'}}), 
	\label{eq:query_update2}
	\end{equation}
	where $\beta$ is the meta learning-rate.
	Note that $\nabla_{(\bm{\theta}, \bm{\alpha})}\emph{L}_{q(\tau_k)}(\bm{\theta^{'}})$ uses the meta-learner's loss on the query set to compute the gradients of $\bm{\theta}$ and $\bm{\alpha}$ but not $\bm{\theta^{'}}$.
	
	By training the meta-learner on lots of FSIC tasks, the meta-learner is forced to learn: 1) the easy fine-tuning initial weights $\bm{\theta}$ for solving FSIC tasks, 2) proper layer-wise adaptive updating learning-rates $\bm{\alpha}$ to benefit the meta-learner's learning from few images.
	With the learned $\bm{\theta}$ and $\bm{\alpha}$, the meta-learner learns on the support set more exactly than other meta-learners which only learn the weight $\bm{\theta}$.
	Algorithm.~\ref{algorithm:train} shows the summarized training procedure of LWAU.

	\section{Experiments}
	\label{sec:experiments}
	\subsection{Dataset}
	We test LWAU on two few-shot classification benchmarks: Miniimagenet\cite{miniimagenet} and Tieredimagenet\cite{tieredimagenet}.
	
	\noindent \textbf{Miniimagenet}\cite{miniimagenet} is a subset sampled from ImageNet\cite{imagenet}.
	It contains 100 image classes, including 64 classes for training, 16 for validation, and 20 for testing.
	Each class in Miniimagenet is composed of 600 images and each image is resized into 84x84 resolution.
	
	\noindent \textbf{Tieredimagenet}\cite{tieredimagenet} is another subset sampled from ImageNet.
	Different from Miniimagenet which contains similar image categories between the training and testing sets (\emph{i.e.} ``pipe organ" in the training and ``electric guitar" in the testing set), Tieredimagenet hierarchically structures the image classes to make all image classes in the testing set are distinct from all classes in the training set.
	It contains 34 high-level image classes, including 20, 6, and 8 classes for training, validation, and testing, respectively.
	Each high-level class is composed of 10 to 30 low-level classes and each low-level class consists of about 1300 images.
	Same to Miniimagenet, all images in Tieredimagenet are resized into 84x84 resolution.
	
	\subsection{Experiment on Miniimagenet}
	We use Conv-4 as the network of the meta-learner and set the number of filters of each convolution layer to 32.
	Each convolution layer is followed with a batch-normalization and a ReLU operator.
	On the training set, we generate 200,000 5-way $K$-shot classification training tasks, and on each of the validation and testing sets, we generate 600 5-way $K$-shot validation or testing tasks.
	$K$ is set to 1 or 5, and in each task, the query set contains 15 samples for each way.
	We train the meta-learner on the training tasks for 60,000 iterations and set the meta batch-size to 4 and the meta learning-rate $\beta$ to 0.001.
	The optimizer we used is Adam\cite{adam}.
	All the updating learning-rates in the vector $\bm{\alpha}$ are initialized to 0.01.
	On each training task, the meta-learner updates itself on the support set for 5 steps, and on each testing task, the meta-learner updates for 10 steps.
	
	Experimental results on Miniimagenet are shown in Tab.\ref{tab:result on Miniimagenet}. 
	Note that, for a fair comparison, all compared methods shown in Tab.\ref{tab:result on Miniimagenet} use Conv-4 as their network.
	The proposed LWAU apparently outperforms the other methods.
	For example, compared with MAML and LLAMA, LWAU promotes the 5-way 1-shot performance for about 2.5\% and 1.1\%, respectively. 

	\begin{table}[t!]
		\centering
		\caption{
			Few-shot Learning Performance on Miniimagenet. 
			The Code of TAML Has not been Released, So We Reimplemented It and Try Our Best to Tune Hyper-parameters for Its Best Performance.
		}
		\resizebox{0.9\columnwidth}{!}{
			\begin{tabular}{|c|c|c|}
				\hline
				\multirow{2}{*}{Method}  &\multicolumn{2}{c|}{5-way Accuracy} \\
				\cline{2-3} &1-shot &5-shot \\
				\hline
				Matching nets FCE\cite{match-network} &44.20\% &57.00\% \\
				\hline
				Meta-learner LSTM\cite{miniimagenet} &43.44$\pm$0.77\% &60.60$\pm$0.71\% \\
				\hline
				MAML\cite{MAML}     &48.70$\pm$1.84\% &63.11$\pm$0.92\% \\
				\hline
				LLAMA\cite{LLAML}    & 49.40$\pm$1.83\% & / \\
				\hline
				iMAML HF\cite{iMAML}  &49.30$\pm$1.88\%	&/ \\
				\hline
				TAML\cite{TAML}  &49.37$\pm$1.79\%	&64.18$\pm$0.82\% \\
				\hline
				LWAU(ours)	&{\bfseries {\color{black} 49.93$\pm$0.83\%}}	&{ \bfseries {\color{black} 64.85$\pm$0.67\%} } \\
				\hline
			\end{tabular}
		}
		\label{tab:result on Miniimagenet}
	\end{table}
	
	\begin{table}
		\centering
		\caption{Few-shot Classification Performance on Tieredimagenet. 
		}
		\resizebox{0.7\columnwidth}{!}{
			\begin{tabular}{|c|c|c|}
				\hline
				\multirow{2}{*}{Method} &\multicolumn{2}{c|}{5-way Accuracy} \\
				\cline{2-3}  &1-shot &5-shot \\
				\hline
				MAML\cite{MAML}              &50.06$\pm$0.85\% &67.95$\pm$0.76\% \\
				\hline
				TAML\cite{TAML}              &50.35$\pm$0.87\% &67.60$\pm$0.71\% \\
				\hline
				LWAU(ours)	&{\bfseries {\color{black} 50.76$\pm$0.89\%}}	&{\bfseries {\color{black} 68.81$\pm$0.68\%}} \\
				\hline
			\end{tabular}
		}
		\label{tab:result on Tieredimagenet}
	\end{table}

	\begin{figure*}[t]
		\centering
		\includegraphics[width=0.85\textwidth]{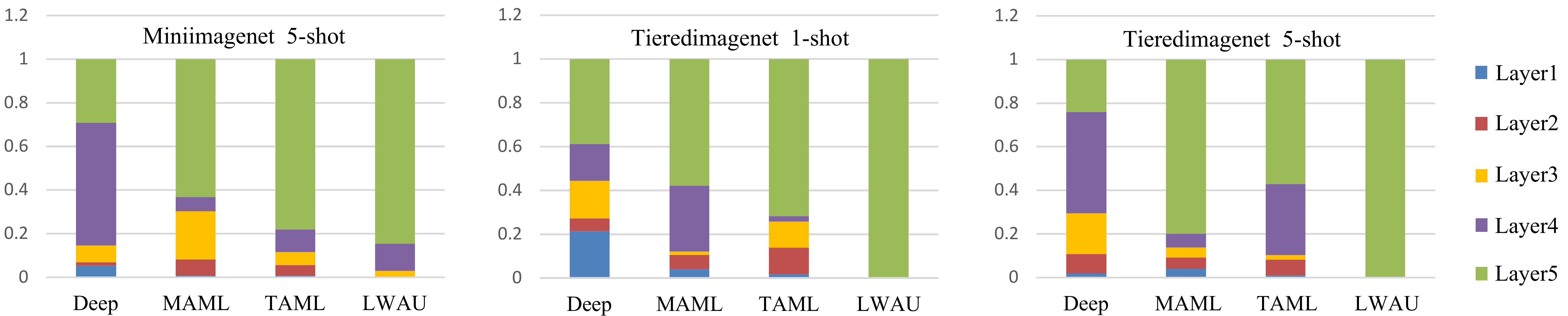}
		\caption{
			The update proportion of each layer in deep model, MAML, TAML, and LWAU, respectively.
			The update proportion is calculated with Eq.\ref{eq:proportion} when they are fine-tuned on the support sets of 600 5-way 1- or 5-shot testing tasks.
			The update proportions on 5-way 1-shot testing tasks on Miniimagenet has been shown before in Fig.\ref{fig:intro}.
		}
		\label{fig:visual}
	\end{figure*}

	\begin{figure}[t]
		\centering
		\includegraphics[width=0.4\textwidth]{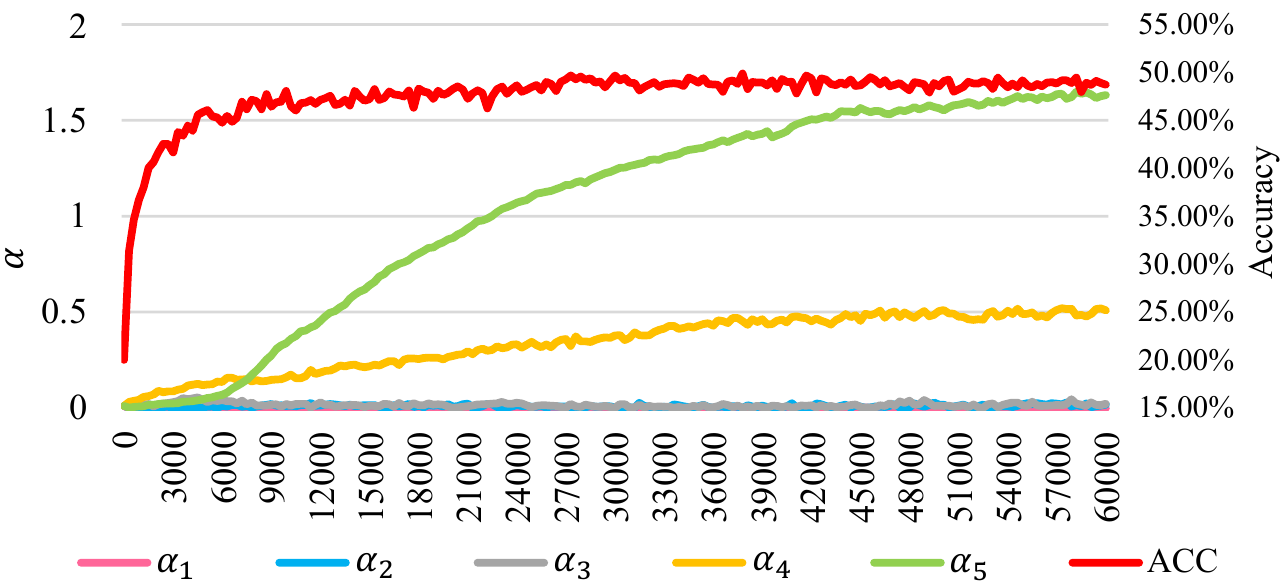}
		\caption{
			The learning curves of $\bm{\alpha}$ and the accuracy.
		}
		\label{fig:visual_curve}
	\end{figure}

	\subsection{Experiment on Tieredimagenet}
	As Tieredimagenet is a larger dataset than Miniimagenet, we set the number of filters of each convolution layer to 64 and generate 400,000 training tasks on the training set.
	We train the meta-learner for 120,000 iterations.
	L1 normalization of 0.001 is applied to prevent the meta-learner from over-fitting, and the meta learning-rate $\beta$ is decreased by 0.5$\times$ for every 10,000 iterations.
	All the other experimental settings are the same as those on Miniimagenet.
	
	Tab.\ref{tab:result on Tieredimagenet} shows the experimental result on Tieredimagenet. 
	Compared with MAML and TAML, LWAU promotes the 5-way 1-shot performance for about 1.4\% and 0.8\%, respectively. 
	Both the experiments on Miniimagenet and Tieredimagenet demonstrate the advantage of the proposed LWAU for the FSIC problem.

	\subsection{Visualization and Analysis}
	\noindent \textbf{Update Proportion} \quad
	To validate the effect of the layer-wise adaptive updating rule, we visualize the update proportion of each layer in Fig.\ref{fig:intro} and Fig.\ref{fig:visual}.
	The calculation of each layer's update proportion has been shown as Eq.\ref{eq:proportion}.
	Compared with the deep model and the other meta-learners, the proposed LWAU meta-learner learns to pay much more attention to update its top layer, especially on the 1-shot learning tasks.
	For example, when solving 5-way 1-shot testing tasks on Miniimagenet and Tieredimagenet, the LWAU meta-learner almost ignores its bottom layers and pays almost all attention to update its top layer (\emph{i.e.}, the update proportion of the 5-th layer is nearly 100\%).
	The visualization supports our assumption that in the FSIC scene, the meta-learner greatly prefers updating its top layer to updating its bottom layers.

	\noindent \textbf{Learning Curve} \quad
	We visualize the learning curves of $\bm{\alpha}$ and the accuracy of LWAU in Fig.\ref{fig:visual_curve}.
	The learning curves are drawn when LWAU is trained on 5-way 1-shot FSIC tasks on Miniimagenet.
	It is clear that $\alpha_4$ and $\alpha_5$ rise up with the training of the meta-learner, while $\alpha_1$, $\alpha_2$ and $\alpha_3$ keep small to near zero across the training.
	LWAU achieves its maximum accuracy of 49.93\% at around the 37,000 iteration.
	
\begin{figure}[]
		\centering
		\includegraphics[width=0.49\textwidth]{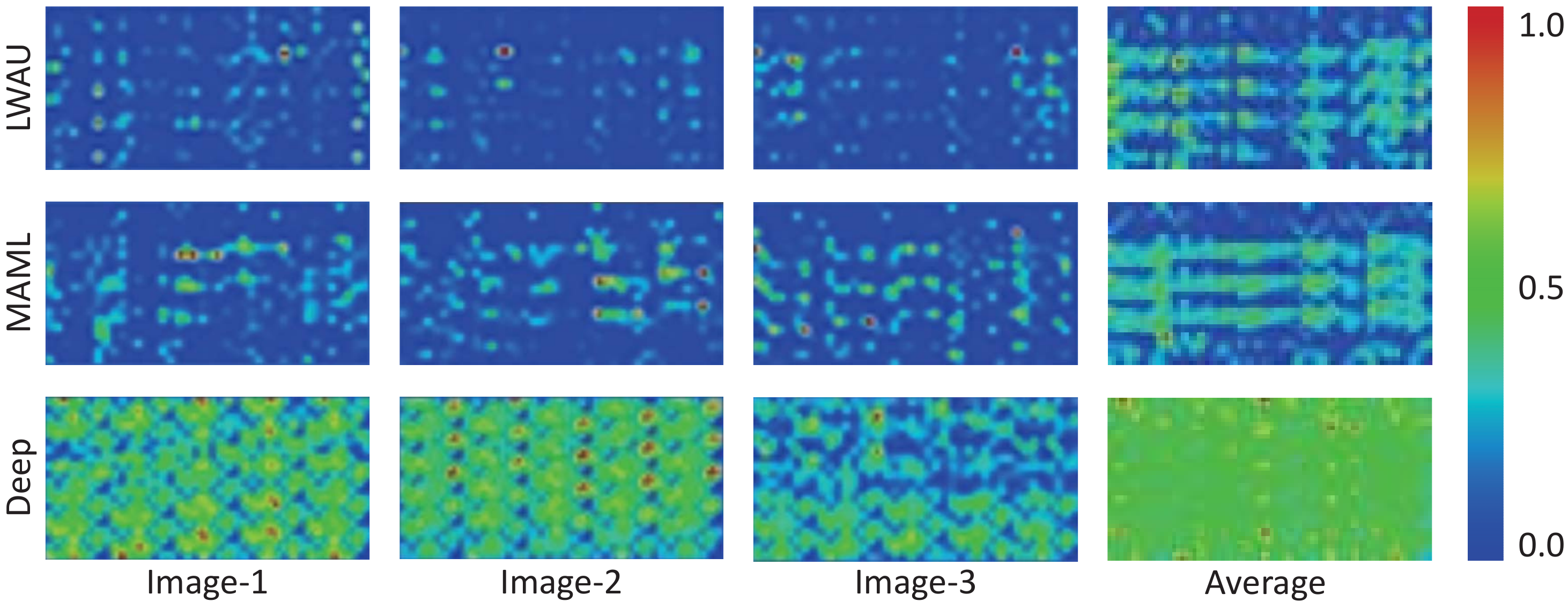}
		\caption{
			The image representations extracted by LWAU, MAML, and deep model.
			All representations are extracted when LWAU, MAML, and deep model are tested on 5-way 1-shot classification tasks on Miniimagenet.
			For better visualization, we reshape each representation vector into a representation map with 20x40 resolution.
			Image-1, 2, 3 are representation maps of three random images from the query set of a testing task.
			Average is the averaged representation map of the images from all 600 testing tasks.
			Each representation map is normalized to have a maximum value of one.
		}
		\label{fig:feature}
	\end{figure}

	\noindent \textbf{Image Representation} \quad
	For a better understanding of LWAU, we visualize its learned image representation in Fig.\ref{fig:feature}.
	The representation which is extracted for the fully-connect layer classifying the input image is an 800 length vector and we reshape the vector into a representation map with 20x40 resolution.
	The representations of MAML and deep model are also shown as comparison and all representations are normalized to have a maximum value of one.
	From Fig.\ref{fig:feature}, we can see that LWAU extracts the sparsest image representation and deep model extracts the densest representation.
	
	We quantify the representation sparsity with the neuron activation percentage.
	The neuron activation percentage of LWAU's representation is about 24.3\%, and those of MAML and deep model are about 30.8\% and 63.1\%, respectively.
	Note that activated neuron is the neuron with a non-zero value that responses to the input image.
	Lots of work\cite{wright2010sparse,wright2009robust,mairal2008discriminative,zhang2017two} have demonstrated that sparse representation benefits image classification and lots of methods\cite{6104179,lu2013face,5597469} improve image classification via utilizing sparse representation.
	Fig.\ref{fig:feature} clearly shows that compared with MAML, the proposed LWAU extracts sparser representation, which might be the reason why LWAU outperforms MAML.

	\subsection{Update Efficiency}
	Fig.\ref{fig:intro} and Fig.\ref{fig:visual} show that the LWAU meta-learner pays little attention to update its bottom layers.
	This indicates that the bottom layers' update might contribute little to LWAU's few-shot classification performance.
	In other words, it is possible to accelerate the LWAU meta-learner's update under the premise of no performance decline via updating only its top layers.
	To verify this point, we do an experiment on Miniimagenet that when tested on an FSIC task, the meta-learner updates only its top layers on support set and freezes its bottom layers.
	
\begin{figure}[]
		\centering
		\includegraphics[width=0.4\textwidth]{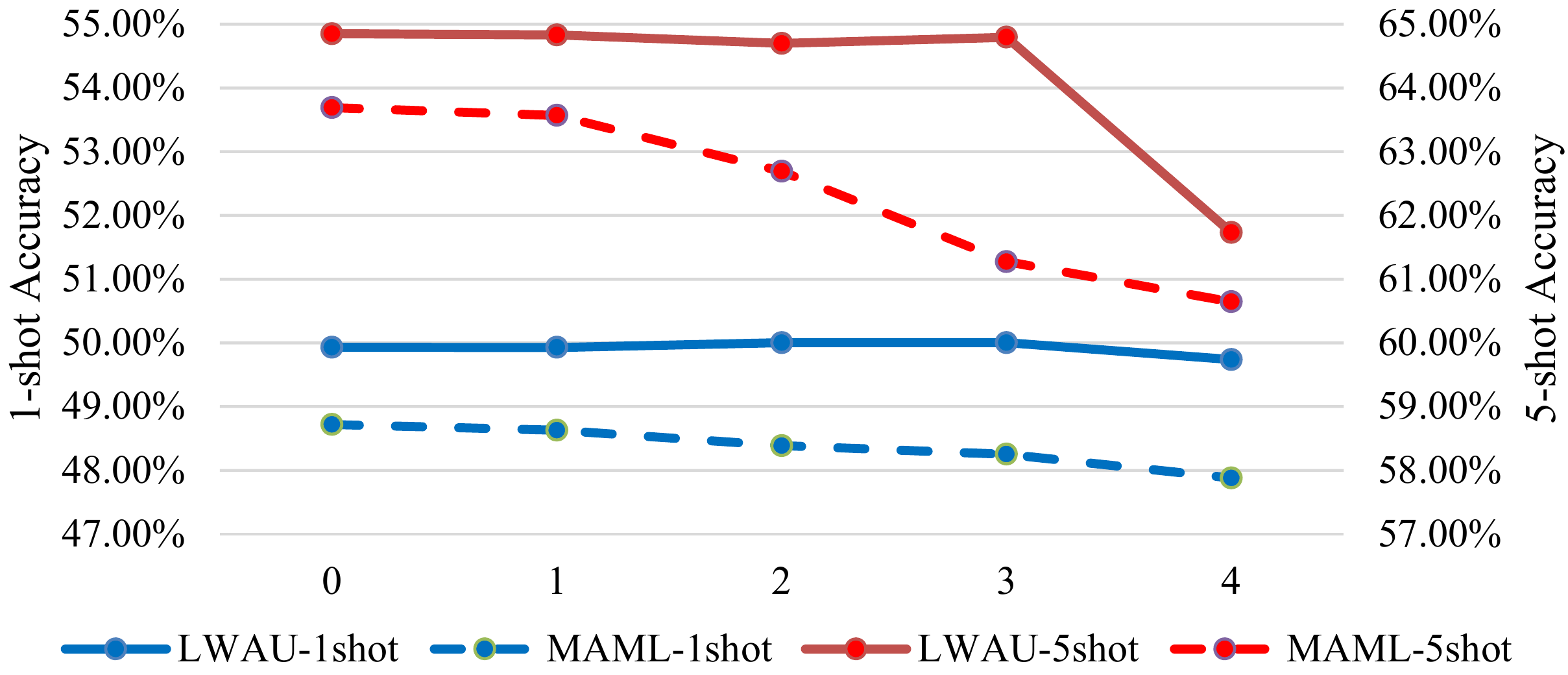}
		\caption{
			Performances of the LWAU and MAML meta-learners when their partial bottom layers are frozen in the update process.
			The left and right $y$-axes denote 1- and 5-shot learning accuracies, respectively.
			Each number $x$ at the $x$-axis denotes the meta-learner's bottom $x$ layers are frozen when inner-updating (\emph{i.e.}, when $x$=0, the meta-learner can update all its layers and when $x$=3, its bottom 3 layers are frozen to avoid update).
		}
		\label{fig:fixed}
	\end{figure}

	The experimental results are shown in Fig.\ref{fig:fixed}.
	Each number $x$ at the $x$-axis denotes the meta-learner's bottom $x$ layers are frozen when it updating on support set.
	When $x$=0, the meta-learner can update all its layers when testing.
	We treat the meta-learner's performance at $x$=0 as its baseline.
	Obviously, Fig.\ref{fig:fixed} shows that freezing the bottom layers effects the LWAU meta-learner hardly while effects MAML greatly.
	For example, when 0$<\!x\!\leq$3, the LWAU meta-learner's performance approximately equivalent to its baseline.
	Whereas, when $x\!\!>$0, the MAML meta-learner performs apparently worse than its baseline.
	This experiment reveals a notable advantage of the LWAU meta-learner that it needs updating only its top layers, which can significantly accelerate its update.
	When $x$=3, the meta-learner's update costs about 6.3 and 11$ms$ on 5-way 1- and 5-shot task, respectively, while the baseline costs 35 and 120$ms$ (we evaluate the meta-learner's update time consuming on one GTX1060 GPU).
	As a conclusion, when tested on an FSIC task, the proposed LWAU can improve its efficiency of learning from few images for at least 5 times by updating only its top 2 layers without sacrificing its FSIC performance.
	
	\section{Conclusion}
	In this paper, we propose a novel meta-learning based layer-wise adaptive updating (LWAU) method to solve few-shot image classification.
	Different from other meta-learning based methods which commonly train a meta-learner to learn only easy fine-tuning weight, LWAU trains a meta-learner to learn not only the easy fine-tuning weight but also a layer-wise adaptive updating rule to improve the meta-learner's learning on few images.
	Extensive experiments show that compared with the other meta-learning based methods, the proposed LWAU achieves not only better few-shot image classification performance but also higher fine-tune efficiency on few images.
	Besides, the visualization of extracted image representations shows that LWAU extracts sparse representations, which benefits to the understanding of LWAU.
	
	{\small
		\bibliographystyle{ieee}
		\bibliography{main}
	}
	

\end{document}